\documentclass[runningheads]{llncs}
\pdfoutput=1 
\usepackage{hyperref}
\hypersetup{
    colorlinks=true,
    linkcolor=blue,
    filecolor=magenta,      
    urlcolor=cyan,
    pdftitle={Overleaf Example},
    pdfpagemode=FullScreen,
    }
\usepackage{comment}
\usepackage[symbol*]{footmisc}
\usepackage{graphicx}
\usepackage[ruled,vlined]{algorithm2e}
\usepackage{booktabs}
\usepackage{amsmath,amssymb}
\usepackage{comment}

\DeclareMathOperator*{\argmin}{arg\,min}

\usepackage{multirow}
\usepackage{graphicx}
\usepackage[table,xcdraw]{xcolor}
\begin{document}
\title{Efficient Global Neural Architecture Search}
\titlerunning{Efficient Global Neural Architecture Search}
%
\author{Shahid Siddiqui\thanks{Corresponding author: msiddi01@ucy.ac.cy}\inst{1}\orcidID{0000-0002-6176-2952}and
Christos Kyrkou\inst{2}\orcidID{0000-0002-7926-7642} \and
Theocharis Theocharides\inst{1,2}\orcidID{0000-0001-7222-9152}
}
\authorrunning{S. Siddiqui et al.}
%
\institute{Electrical and Computer Engineering Department, University of Cyprus, Nicosia, Republic of Cyprus \and
KIOS Research and Innovation Center of Excellence, University of Cyprus, Nicosia, Republic of Cyprus
\\
\url{https://www.kios.ucy.ac.cy/} \\
\email{\{msiddi01,kyrkou.christos,ttheocharides\}@ucy.ac.cy}}
\maketitle              
\begin{abstract}

Neural architecture search (NAS) has shown promise towards automating neural network design for a given task, but it is computationally demanding due to training costs associated with evaluating a large number of architectures to find the optimal one. To speed up NAS, recent works limit the search to network building blocks (modular search) instead of searching the entire architecture (global search), approximate candidates' performance evaluation in lieu of complete training, and use gradient descent rather than naturally suitable discrete optimization approaches. However, modular search does not determine network's macro architecture i.e. depth and width, demanding manual trial and error post-search, hence lacking automation. In this work, we revisit NAS and design a navigable, yet architecturally diverse, macro-micro search space. In addition, to determine relative rankings of candidates, existing methods employ consistent approximations across entire search spaces, whereas different networks may not be fairly comparable under one training protocol. Hence, we propose an architecture-aware approximation with variable training schemes for different networks. Moreover, we develop an efficient search strategy by disjoining macro-micro network design that yields competitive architectures in terms of both accuracy and size. Our proposed framework achieves a new state-of-the-art on EMNIST and KMNIST, while being highly competitive on the CIFAR-10, CIFAR-100, and FashionMNIST datasets and being 2-4$\times$ faster than the fastest global search methods. Lastly, we demonstrate the transferability of our framework to real-world computer vision problems by discovering competitive architectures for face recognition applications.

\footnote{Code and results available at: \href{https://github.com/siddikui/Efficient-Macro-Micro-NAS}{https://github.com/siddikui/Efficient-Macro-Micro-NAS}}


\end{abstract}

\section{Introduction}
\label{intro}

Neural Architecture Search (NAS) is the task of automating the neural network design process for a given dataset. NAS works based on reinforcement learning (RL) and evolution (EA)\cite{zoph2016RL,MetaQNN,Real_evolution,GeneticProgrammingCNN} achieve excellent design automation for computer vision tasks, but require enormous computational resources hindering their widespread adoption. The computational expense of NAS is due to; (1) excessively large network design spaces, often containing billions of candidate architectures, (2) each candidate requiring training (which itself is expensive) to determine how it performs as compared to others, and 3) discrete search strategies attempting to evaluate a large number of candidates. Therefore, NAS research is highly focused on efficiency by adopting modular search instead of global \cite{NasNet,liu_darts}, and by approximating networks' performance instead of expensive training \cite{EAS,NASH,liu_darts,AGNAS,performancepredictionpower}. For details of NAS research, we refer the reader to \cite{NAS1000Papers}. 


Modular NAS, first introduced by NASNet \cite{NasNet} leads to efficient search, e.g., from several thousands \cite{zoph2016RL} to 4 GPU-days \cite{liu_darts,AGNAS}. Hence, it has been widely researched by the NAS community, but has certain drawbacks. Modular NAS proposes to search only for a network module, also called a cell, rather than searching the entire architecture (global NAS). However, once a module is discovered for a given dataset, the total number of modules and channels (network's depth and width, i.e. macro architecture) is still unknown and needs manual intervention. This contradicts the main objective of automatically designing a suitable network for a given dataset. Another fundamental issue of modular search spaces is their narrow accuracy range and even randomly sampled architectures perform well \cite{frustrating}. Such spaces may yield good results quickly, but do not possess architectural diversity suitable for a variety of datasets. Therefore, in \cite{CAIP23} we propose a flexible search space in terms of accuracy and network complexity. In this work, we further show how the design principles introduced in \cite{CAIP23}, can easily be adapted for real-world applications. More specifically, we expand on the concepts to develop ResNet \cite{ResNet} based search space for face recognition applications that offers more accurate and efficient networks than those used by leading face recognition methods \cite{Adaface,Arcface}. The same design principles can be used for other computer vision applications.

To establish relative rankings of networks, early methods \cite{zoph2016RL} train and evaluate a large number of candidates (12,800 networks each for 50 epoch), which is extremely computationally intensive. Many subsequent works propose performance approximations instead of training each network from scratch. \cite{EAS,NASH} suggest reusing the weights of already trained networks during the search, but it becomes unclear whether the accuracy gain is due to the pre-trained weights or the superiority of the discovered architecture. Similarly, techniques based on parameter sharing in the supernet \cite{liu_darts,AGNAS} lead to inaccurate relative rankings as discussed in \cite{EvaluatingSearch}. Many performance predictors have also been proposed\cite{performancepredictionpower}, but training and evaluating predictors itself requires ground-truth validation accuracies of a large number of trained candidates, which is impractical to acquire for every new use case. To address this problem, we propose using a true ranking of the architectures in \cite{CAIP23}. However, another issue is that all networks are trained under one training protocol, while training hyperparameter settings has been shown to have a greater effect on accuracy than the architecture itself \cite{frustrating,JAHS}. Keeping that in mind, we propose a novel ranking mechanism that assigns the training protocol relative to the learning capacity of the candidates, leading to a high correlation with their final performance compared to when they are all trained under similar settings. This allows a fairer and faster comparison of candidates, hence guiding the search strategy accurately and efficiently. 

In terms of search strategy, NAS works treat the entire search space as one huge combinatorial space, therefore naturally apply discrete optimization algorithms \cite{zoph2016RL,MetaQNN,Real_evolution,NasNet}. This usually necessitates evaluation of a large number of networks, hence a surge in the search cost. In this work, we split the search into macro and micro architecture discovery. In addition to reducing search costs, this allows the algorithm to first discover a good trade-off between network complexity and the given dataset difficulty. Once a good outer skeleton is learned, micro search can further enhance the fine grain-architecture. In summary, our main contributions are the following:

\noindent• A minimal but architecturally diverse search space.

\noindent• A more accurate and faster network ranking mechanism.  

\noindent• An efficient search algorithm for end-to-end network discovery.

\noindent• Transferability of our framework's design principles to real-world computer vision applications. 

Our proposed framework discovers competitive architectures in terms of both accuracy and size for CIFAR-10, CIFAR-100 \cite{CIFAR} and FasionMNIST \cite{Fashion} while being 2$\times$ to 4$\times$ faster than the fastest global NAS method \cite{NASH}. Moreover, compared to the best manually designed networks, we achieve new state-of-the-art results on EMNIST \cite{EMNIST} and KMNIST \cite{KMNIST}. For real-world applications of 1:1 face verification and top-n identification on a challenging Tinyface \cite{TinyFace} dataset, we perform better than a leading face recognition method Adaface \cite{Adaface} with networks up to 2$\times$ smaller than commonly used ResNets\cite{ResNet}, therefore bridging the gap between theory and applications.


\section{Related Work}

In this work, we reemphasize the automation of the network design process, therefore our work is closely related to the first work \cite{Macro} that revisits NAS for end-to-end discovery with minimal human intervention. This line of work is further emphasized by the more recent work of \cite{Macro-Neural}, which proposes automatically generated search spaces from existing architectures. Another recent work is that of \cite{AGNAS} as it focuses on both micro and macro search; however, it still manually stacks up the blocks to decide the final architecture. Further, we divide the related work into two parts; (1) neural architecture search and (2) face recognition applications.

\subsection{Neural Architecture Search}

\subsubsection{Search Space} End-to-end NAS methods search for network depth, width, and convolutional kernel size \cite{zoph2016RL,MetaQNN,Real_evolution,NASH,EAS,NASBOT} except \cite{GeneticProgrammingCNN}. In addition, all these works search for skip connections except \cite{MetaQNN,EAS}. Some works follow the VGG-like CONV-POOL-FC design paradigm \cite{MetaQNN,EAS,NASBOT}. We analyze these existing works in depth in Section \ref{SSDesign} and propose search space optimizations.

\subsubsection{Search Strategy}

The algorithms most commonly used by the research community are based on reinforcement learning (RL) \cite{zoph2016RL,MetaQNN,EAS}, evolutionary algorithms \cite{Real_evolution,GeneticProgrammingCNN,NASH,Macro-Neural}, and sequential model-based optimization (SMBO) \cite{NASBOT,NSGA} methods. Due to the huge search spaces, it is computationally expensive for these methods to accurately evaluate a large number of candidates. In Section \ref{OurAlgorithm}, we propose an algorithm that can discover architectures efficiently.

\subsubsection{Performance Evaluation}

Early NAS works \cite{zoph2016RL,MetaQNN,Real_evolution,GeneticProgrammingCNN} fully train candidates to establish their relative ranking and employ speed-up strategies, but an excessively large number of architectures to be evaluated leads to enormous compute costs. Hence, a significant body of NAS research lies in the speed of candidate evaluation. Model-based accuracy predictors are fairly simple \cite{modelbasedpredictor} but require architecture-accuracy paired data for every new dataset, which itself is expensive to obtain, and hence are not generalizable. \cite{EarlyTermination} and \cite{MDENAS} use learning curve prediction, and estimate the final rankings by training for fewer epochs, respectively. Although these methods may perform better for modular search spaces, our experiments in a global space show low correlation between early and final performance of the networks. The most widely adopted technique is perhaps parameter sharing in a supernet \cite{ENAS}, where subgraphs can inherit weights from a large over-parameterized network. \cite{EvaluatingSearch} shows that weight sharing leads to inaccurate rankings. Recently, there has been a substantial body of research using zero-cost proxies \cite{zerocostproxies}, which require no training or very little training. However, works such as \cite{antizeroshot,antizeroshotNIPS} show that proxies developed for modular search spaces are not transferable to global spaces and existing zero-shot proxies cannot outperform simple baseline such as number of parameters and FLOPS. In general, excessive focus on modular search has lead to a substantial research gap for efficient evaluation methods for global NAS.

\subsection{NAS for Face Recognition}

To demonstrate the transferability of our framework to face recognition applications, we use ResNets \cite{ResNet}, as modified in \cite{Adaface,Arcface}, as our baseline architectures. \cite{FairFace} applies joint NAS and hyperparameter optimization (HPO) to mitigate bias in face recognition. \cite{TeachNAS,PocketNet} combine knowledge distillation and NAS to achieve low-complexity, high-accuracy networks. However, our method discovers high-accuracy low-complexity networks solely based on the NAS framework. Although relevant, these methods are highly specialized for face recognition and hence are not directly comparable to our NAS only method.    

\section{Methodology}

In this section, we discuss our proposed search space, search strategy, and performance evaluation method.  

\subsection{Search Space Design}
\label{SSDesign}

A \textbf{search space}, or just \textbf{space} from here on, is defined as a set of network variables from which various architectural configurations can be sampled. Table \ref{tab:searchspace} shows that most existing spaces \cite{zoph2016RL,MetaQNN,EAS,NASBOT} are influenced by the early Conv-Pool-FC-like architecture paradigm \cite{VGG} and/or residual networks \cite{ResNet}. Moreover, network depth, width, and kernel size are the most common variables followed by convolution stride (Strides), skip connections, pooling layers, and fully connected (FC) layers. Since the number of network configurations grows exponentially with the number of search variables, we set the variables so that the resulting space is navigable for our search strategy. However, it is not sufficient to just drop most of the variables for efficiency gains since this may limit the space in terms of performance. Hence, we strike a balance between automation and search space effectiveness, i.e. wide range performance/complexity trade-off. Such a space can better adapt to varying complexity tasks, by offering smaller networks for easier tasks and relatively complex networks for harder ones. Next, we discuss the optimizations done to create such a search space.

\begin{table}[t]
\centering
\caption{Search Space Comparison: The proposed search space focuses on the most impactful design choices in terms of efficiency.}
\label{tab:searchspace}
\resizebox{\textwidth}{!}{%
\begin{tabular}{@{}lcccccccc@{}}
\toprule
\textbf{NAS Method}                                   & \multicolumn{8}{c}{\textbf{Global Search Space Architectural Variables}} \\ \midrule
\multicolumn{1}{c|}{\textbf{}} &
  \textbf{\begin{tabular}[c]{@{}c@{}}Depth\\ (Layers)\end{tabular}} &
  \textbf{\begin{tabular}[c]{@{}c@{}}Width\\ (Channels)\end{tabular}} &
  \textbf{\begin{tabular}[c]{@{}c@{}}Operations\\ per Layer\end{tabular}} &
  \textbf{\begin{tabular}[c]{@{}c@{}}Convolutional\\ Kernel\end{tabular}} &
  \textbf{Strides} &
  \textbf{\begin{tabular}[c]{@{}c@{}}Pooling\\ Layers\end{tabular}} &
  \textbf{\begin{tabular}[c]{@{}c@{}}Fully\\ Connected\\ Layers\end{tabular}} &
  \textbf{\begin{tabular}[c]{@{}c@{}}Skip\\ Connections\end{tabular}} \\ \midrule
\multicolumn{1}{l|}{\textbf{NAS-RL\cite{zoph2016RL}}}                  & \checkmark       & \checkmark       &         & \checkmark      & \checkmark      & \checkmark      &        & \checkmark      \\
\multicolumn{1}{l|}{\textbf{Meta-QNN\cite{MetaQNN}}}                & \checkmark       & \checkmark       &         & \checkmark      & \checkmark      & \checkmark      & \checkmark      &        \\
\multicolumn{1}{l|}{\textbf{Large-scale Evolution\cite{Real_evolution}}}   & \checkmark       & \checkmark       &         & \checkmark      & \checkmark      &        &        & \checkmark      \\
\multicolumn{1}{l|}{\textbf{EAS\cite{EAS}}}                     & \checkmark       & \checkmark       &         & \checkmark      & \checkmark      & \checkmark      & \checkmark      &        \\
\multicolumn{1}{l|}{\textbf{Genetic Programming CNN\cite{GeneticProgrammingCNN}}} &         & \checkmark       &         & \checkmark      &        &        &        & \checkmark      \\
\multicolumn{1}{l|}{\textbf{NASH-Net\cite{NASH}}}                & \checkmark       & \checkmark       &         & \checkmark      &        &        &        & \checkmark      \\
\multicolumn{1}{l|}{\textbf{NASBOT\cite{NASBOT}}}                  & \checkmark       & \checkmark       &         & \checkmark      & \checkmark      & \checkmark      & \checkmark      & \checkmark      \\
\multicolumn{1}{l|}{\textbf{TRG-NAS\cite{CAIP23}}}               & \checkmark       & \checkmark       & \checkmark       & \checkmark      &        &        &        &        \\ \bottomrule
\end{tabular}%
}
\end{table}

\subsubsection{Trimming Search Variables} To start with, we can drop variables arising from early Conv-Pool-FC-like architectures \cite{VGG} by leveraging FCN-like networks \cite{FCN}. Therefore, FC layers can be replaced by a global pooling layer, and pooling layers can be replaced by convolutions with stride 2. Additionally, we can drop skip connections since we are not explicitly seeking very deep networks. Hence, we trim down \textbf{Fully connected}, \textbf{Pooling layers}, and \textbf{Skip connections} from Table \ref{tab:searchspace}. 
  
\subsubsection{Channels Search Reduction} Table \ref{tab:searchspace} shows that all methods search for width (number of channels). This is usually done for each layer as in \cite{zoph2016RL}. However, we limit the search for the channel to the initial layer only and use a fixed rate of channel doubling whenever the spatial dimensions are halved, as in \cite{VGG,liu_darts}. Additionally, it allows us to fix stride values of convolutions to 2 whenever doubling the channels and 1 otherwise. Hence, we drop \textbf{ strides} from the search space. This technique further reduces the search complexity (discussed in the next section), but still allows variable-width architectural diversity.  

\subsubsection{Performance/Complexity Trade-off} We notice that existing global spaces do not allow operation search whereas operation type such as separable, dilated or plain convolution can allow significant expressiveness. Although we target a compact space, it should still maintain the idea of previously unseen and novel architectures. Hence, we allow searching for \textbf{Operation} type as either separable or plain convolution. For every layer, operation choice coupled with kernel choice of 3 or 5 creates a good number of possible micro architecture variations.

\smallskip
To this end, we propose a search space with \textbf{\textit{depth}}, \textbf{\textit{width}}, \textbf{\textit{operations}}, and \textbf{\textit{kernels}} variables, as shown in Table \ref{tab:searchspace}. To study the accuracy variance of this space, we randomly sample 240 networks and train each network for 50 epochs on CIFAR-10. The worst network achieves an accuracy of 76.11\%, while the best network is 94.65\%. For comparison, the most widely adopted DARTS space has its worst network achieving 96.18\% and the best 97.56\% from within 214 sampled architectures by \cite{frustrating}. This shows that our search space has a high variance in terms of performance, therefore, better discovered networks can be attributed to the superiority of the search algorithm and not to expertly crafted space.

\subsubsection{Search Space Complexity}

The complexity of the space depends on search bounds and increases exponentially with depth. For a depth range of $D$, a width range of $W$, a number of operations $O$, a number of kernels $K$, and a final discovered depth $D_{f}$, the maximum number of architectures $N_{arch}$ is given in the equation. \ref{eq:compl}. 

\begin{equation}
N_{arch} = (O  \times  K)^{D_{f}} \times D \times W
\label{eq:compl}
\end{equation}

We limit the depth range between 5 to 100 layers, channels from 16 to 128 with steps of 2, two operations and two kernel sizes i.e., $D = 96$ and $W = 57$, $O = 2$ and $K = 2$. Then, assuming $D_{f} = 25$, the space as described above has approximately $6.16 \times 10^{18}$ candidate architectures. For comparison, the search space for darts \cite{liu_darts} is also approximately $10^{18}$ but has a much narrower accuracy range than ours. With the search variables explained, we can now formally define the search problem.

\subsubsection{Search Problem}
Let $\mathcal{L}_{train}$ and $\mathcal{L}_{test}$ denote training and test loss, respectively. These are determined by the network architecture $x$ and its weights $\theta$.
The search goal is to find $x^*$ that minimizes the test loss $\mathcal{L}_{test}(\theta^*,x^*)$, where the weights $\theta^*$ associated with the architecture are obtained by minimizing the training loss $\theta^* = \argmin_{\theta} \mathcal{L}_{train}(\theta, x^*)$. This is a bilevel optimization problem with $x$ as outer-level optimization variables and $\theta$ as inner-level optimization variables:

\begin{equation}
\min_{x \in \mathcal{X}} \mathcal{L}_{test}(\theta^*(x),x)
\label{eq:opt1}
\end{equation}

\begin{equation}
\textbf{s.t.    } 
\theta^*(x) = \argmin_{\theta} \mathcal{L}_{train}(\theta, x)
\label{eq:opt2}
\end{equation}

where $\mathcal{X} = \lbrace (D, W, O, K) \mid D \in [D_{\min}, D_{\max}], W \in \lbrace W_{\min} + ne \mid n \in \mathbb{N}_0, e \in E \rbrace, 
O \in \lbrace o_1, o_2 \rbrace, K \in \lbrace k_1, k_2 \rbrace \rbrace$. $D$, $W$, $O$ and $K$ determine network depth, width, operation type, and kernel size, respectively.

\subsection{Performance Evaluation}
\label{perfeval}

Solving Equation \ref{eq:opt2} by training each candidate from scratch and until convergence is expensive, hence previous work has used various approximation methods \cite{EAS,NASH,liu_darts,performancepredictionpower}. In the development of these methods, it is assumed that all networks are comparable under a single training protocol, while \cite{frustrating,JAHS} suggest that the training protocol may affect the accuracy more than the architecture itself \cite{frustrating,JAHS}. In this section, we show that ranking networks by training them differently (dynamic learning rankings), leads to more accurate relative rankings as compared to when they are all trained under similar settings (static learning rankings).

\subsubsection{Static Learning Rankings} 

We randomly select 240 networks from our search space, train each for 50 epochs on CIFAR-10 following \cite{zoph2016RL}, and record their validation accuracies at every epoch. We calculate Spearman rank correlation metric \cite{Spearman} between `models trained for 1 epoch' and `the same models trained for 50 epochs' i.e. their final performance. We find that the correlation value is \textit{ 0.65}, indicating that \textbf{\textit{rankings of networks determined by shorter training are not as accurate as when fully trained}}. Then, we calculate the same metric between `number of parameters of models' or simply \textbf{params} and their final performance. The correlation value turns out to be \textit{0.49}, showing that \textbf{\textit{ the number of parameters alone are not indicative of the final performance of the network}}, i.e. larger networks are not necessarily better networks.

\subsubsection{Dynamic Learning Rankings} 

Next we sample 50 networks from 240 and sort w.r.t. their parameters in ascending order. For each network increasing in parameters, we record its validation accuracy against training for an additional epoch. For example, the smallest network trained only for 1 epoch, the next one for 2 epochs, and so on. We observe a high correlation value of \textit{0.91} between params and corresponding `validation accuracy with increasing epochs' (dynamic learning rankings). This indicates that \textbf{\textit{a candidate may perform better relative to others due to increased parameters or better training, or both}}. In addition, networks ranked using dynamic learning have a high correlation value of \textit{0.85} with their final performance, ie \textit{ 20\%} better than static learning rankings. Therefore, \textbf{\textit{not all networks need to be trained until convergence and their training protocol can be chosen relative to each other}}.

\subsubsection{Applicability to NAS} 

We use the insights gained from the above experiments and employ a dynamic learning evaluation mechanism to determine relative rankings of candidates during the search as follows: 1) Whenever we evaluate a candidate with more parameters compared to its neighbor, we also train it for an additional epoch, 2) However, while comparing model with lesser parameters compared to its neighbor, we expect to improve performance due to better training, so we train models even longer, 3) Finally, we follow consistent training when comparing models with similar number of parameters. In this way, we are able to achieve better relative ranking without fully training all models, hence a much faster evaluation scheme.

\subsection{Search Algorithm}
\label{OurAlgorithm}

The proposed search space is still large; however, we have set the architectural variables in such a way that we can split the global search into macro and micro searches, enabling our algorithm \ref{algorithmmain} to navigate it efficiently. 

\subsubsection{Macro Architecture Search} 

We initialize the search with a small number of evaluation epochs $E_{min}$, minimum depth $D_{min}$, maximum width $W_{max}$ and all operations and kernel sizes set to separable convolution ($Sep$) and $3\times3$, respectively. We let candidates \textbf{\textit{Grow}} layers in an attempt to achieve better validation accuracy. As discussed in section \ref{perfeval}, for each added layer, the candidates are trained for an additional epoch compared to the previous one. Layers are added until they increase performance by $L^{+}_{acc^+}$(accuracy gain by adding layer). By adding a layer, if the accuracy does not drop more than $L^{+}_{acc^-}$(accuracy drop by adding layer), we continue searching for depth. The depth search is terminated if $D_{max}$ (upper bound of layers) is reached or the accuracy drops more than $L^{+}_{acc^-}$ three times. We empirically determine that the threshold values for $L^{+}_{acc^+}$ and $L^{+}_{acc^-}$ are 0.10 and 0.05, respectively. Once the depth search ends, we \textbf{\textit{Prune}} the number of channels until $W_{min}$ is reached or the accuracy drops below the previous best three times. Since we are decreasing the learnable parameters when pruning, we expect the performance to drop, therefore, for every pruned candidate, we evaluate it with 2 additional training epochs, as mentioned in Section \ref{perfeval}. After the width search ends, we have an architecture with $D_{f}$ and $W_{f}$, i.e., final depth and width, respectively, while $O$ and $K$ remain unchanged. The macro search discovers a good depth and width for the given dataset while reducing the right term of complexity in the equation. \ref{eq:compl} to $D'$+$ W'$, where $D'$ and $W'$ are the number of networks evaluated for depth and width, respectively. 

\label{searchalgorithm}
\begin{algorithm}[t]
\caption{NAS Search Algorithm}
\SetKwInput{KwInput}{Input}                
\SetKwInput{KwVariables}{Initialization} 
\DontPrintSemicolon
\KwInput{Search bounds: $D_{min}$, $D_{max}$, $W_{min}$, $W_{max}$, $W_{res}$ \;}
\KwVariables{
    $E = E_{min}$, $L = D_{min}$, $C = W_{max}$, $O = Sep$, $K = 3\times3$ \;
}
\textit{\textbf{MACRO Search}}\;
1. \textit{\textbf{Grow}} model $L \gets L + 1$ \& \textbf{train} for $E \gets E + 1$ \textit{\textbf{while}} $Acc^{test}$ improves\;
2. \textit{\textbf{Prune}} model $C \gets C - W_{res}$ \& \textbf{train} for $E \gets E + 2$ \textit{\textbf{while}} $Acc^{test}$ retains\;
\textit{return \textbf{Macro} architecture $D_{f}$, $W_{m}$, $O$, $K$}\;

\textit{\textbf{MICRO Search}}\;
$E = E_{min}$, $L = D_{f}$, $C = W_{m}$, $O = Sep$, $K = 3\times3$ \;
1. \textit{Retrain \textbf{Macro}} architecture for $E_{min}$ as new baseline\;
2. \textit{\textbf{Replace}} $O_i \gets Conv$ \& $C \texttt{-{}-}$ \textit{\textbf{if}} $Acc^{test}$ improves by \textbf{training} for $E_{min}$\;
3. \textit{\textbf{Update}} $K_i \gets [5\times5]$ \& $C \texttt{-{}-}$ \textit{\textbf{if}} $Acc^{test}$ improves by \textbf{training} for $E_{min}$\;
\textit{\textbf{Return} global architecture with $D_{f}$, $W_{f}$, $O_{f}$, $K_{f}$}\;
\label{algorithmmain}
\end{algorithm}

\subsubsection{Micro Architecture Search}

Macro search adapts the architecture to a good performance point. We subsequently try to fine-tune it with operation type and kernel size at each layer. However, in addition to the micro space being combinatorially huge, see equation \ref{eq:compl}, plain convolution and larger kernel size drastically increase network parameters. This is problematic in the sense that we would like to achieve better performance as a result of better architectures and not due to increased parameters. Therefore, for micro search, whenever we use plain convolution or larger kernel size, we decrease the number of channels such that the total number of parameters remains approximately equal to that of baseline macro architecture. There are two benefits of this approach, 1) we can compare all micro architectures with an equal and small number of training epochs as discussed in Section \ref{perfeval}, 2) we can evaluate a reasonably high number of micro architectures. Therefore, we initiate micro search by retraining the discovered macro architecture with $E_{min}$ as a baseline and then perform search operations and kernel sizes for each layer. We \textbf{\textit{Replace}} separable convolutions with plain ones and \textbf{\textit{Update}} kernel sizes if the validation accuracy improves by training for $E_{min}$. Micro search ends once all layers are traversed for alternative operations and kernels. Hence, we discover $O_{f}$ and $K_{f}$ by evaluating $2\times D_{f}$ architectures. At this point, we have adapted an architecture for the target dataset by evaluating only $N_{evaluated} = 2\times D_{f}+D'+W'$ architectures instead of the number shown in Eq.\ref{eq:compl}.

\subsubsection{Parameter Efficient Networks} As shown in Algorithm \ref{algorithmmain}, we initialize the search with minimum depth, maximum width, and all layers of separable convolutions with kernel sizes of $3\times3$. This decision is reached by empirically evaluating alternative initialization strategies where layers can initially be convolutions or kernel sizes be $7\times 7$, as shown in Table \ref{tab:alternativeoptim}. To single out the contribution of each strategy and for faster evaluation, we sample 10 binary subdatasets from CIFAR-10 instead of using the entire dataset and record the averaged accuracy and number of parameters. In Table \ref{tab:alternativeoptim}, we show that the best strategy is to start with smaller networks and add parameters only if performance improves. This strategy significantly beats others in terms of the accuracy/parameter efficiency trade-off.

\begin{table}[t]
\small
\centering
\caption{Effect of different initialization strategies on search. }
\label{tab:my-table}

\begin{tabular}{lcccc}
\hline
\textbf{Initialization Strategy}             & \multicolumn{1}{l}{Conv-64-3x3} & \multicolumn{1}{l}{Conv-64-7x7} & \multicolumn{1}{l}{Sep-64-3x3} & \multicolumn{1}{l}{Sep-64-7x7} \\ \hline
\multicolumn{1}{l|}{\textbf{Accuracy (\%)}}  & 97.85                           & 97.35                           & \textbf{97.96}                 & 97.73                          \\
\multicolumn{1}{l|}{\textbf{Parameters (M)}} & 0.65                            & 0.64                            & \textbf{0.23}                  & 0.90                           \\ \hline
\end{tabular}%

\label{tab:alternativeoptim}

\end{table}


\section{Experiments and Results}

This section is divided into two subsections. Section \ref{Classification} is focused on experiments and results for image classification datasets. Section \ref{Recognition} details the transferability of our framework to facial recognition applications.

\subsection{Image Classification Experiments and Results}
\label{Classification}

\subsubsection{Datasets} For comparison with NAS methods, we use the CIFAR-10\cite{CIFAR}, CIFAR-100\cite{CIFAR}, and FashionMNIST\cite{Fashion} datasets. To compare with the previous manually designed state-of-the-art approaches, we use EMNIST balanced\cite{EMNIST} and KMNIST\cite{KMNIST} datasets. 

\subsubsection{Search and Training Settings } \label{ClassificationSet} For all datasets, we perform the search under two settings i.e. \textbf{tiny} and \textbf{mobile} networks. For \textbf{tiny} settings, we run search with $D_{min}$=10, $D_{max}$=100, $W_{min}$=16, $W_{max}$=64, $W_{res}$=2 and $E_{min}$=10 for macro search and $E_{min}$=2 for micro search. For \textbf{mobile} settings, we simply increase the $W_{max}$=128. During the search, we train all candidates with varying epochs (as discussed in our search strategy) using SGD with momentum of 0.9 and weight decay of 3e-4. We used an initial learning rate of 0.025 annealed down to 0 using a cosine scheduler, batch size of 64 and cutout augmentation\cite{cutout}. For final training, we use the same settings as search except epochs, which are fixed at 600. In general, we use standard training settings as in \cite{liu_darts,Macro-Neural} and do not use enhanced training protocols that hide the contributions of the search strategy or the search space \cite{frustrating}. To show the contributions of our proposed method, we follow the best practices of NAS as suggested by \cite{BestNASPractice,frustrating}. All experiments were carried out on a single Nvidia Quadro RTX 8000 GPU.

\begin{table}[t]
\centering
\caption{Comparison with state-of-the-art NAS architectures for CIFAR-10.}
\label{tab:C10}
\resizebox{\textwidth}{!}{
\begin{tabular}{l|ccccc}
\hline
\textbf{NAS Method} &
  \textbf{\begin{tabular}[c]{@{}c@{}}Test Err.\\ (\%)\end{tabular}} &
  \textbf{\begin{tabular}[c]{@{}c@{}}Params\\ (M)\end{tabular}} &
  \textbf{\begin{tabular}[c]{@{}c@{}}Search Cost\\ (GPU-days)\end{tabular}} &
  \textbf{\begin{tabular}[c]{@{}c@{}}Search\\ Space\end{tabular}} &
  \textbf{\begin{tabular}[c]{@{}c@{}}Search\\ Algorithm\end{tabular}} \\ \hline
\textbf{ResNet\cite{ResNet}}                  & 6.43 & 1.7 & - & -  & -       \\ \hline
\textbf{NAS-RL\cite{zoph2016RL}}                  & 3.65 & 37.4 & 22400 & Global  & RL       \\
\textbf{Meta-QNN\cite{MetaQNN}}                & 6.92 & 11.2 & 100   & Global  & RL       \\
\textbf{EAS\cite{EAS}}                     & 4.23 & 23.4 & 10    & Global  & RL       \\
\textbf{Large-scale Evolution\cite{Real_evolution}}   & 5.40 & 5.4  & 2600  & Global  & EA       \\
\textbf{Genetic Programming CNN\cite{GeneticProgrammingCNN}} & 5.98 & 1.7  & 14.9  & Global  & EA       \\
\textbf{NASH-Net\cite{NASH}}                & 5.20 & 19.7 & 1     & Global  & EA       \\
\textbf{Macro-NAS\cite{Macro-Neural}}               & 4.23 & 6.7  & 1.03  & Global  & EA       \\
\textbf{RandGrow\cite{Macro}}                & 3.38 & 3.1  & 6     & Global  & RS       \\
\textbf{Petridish\cite{Petridish}}               & 2.83 & 2.2  & 5     & Global  & Gradient \\
\textbf{NASBOT\cite{NASBOT}}                  & 8.69 & N/A  & 1.7   & Global  & SMBO     \\
\textbf{NSGA-NET\cite{NSGA}}                & 3.85 & 3.3  & 8     & Global  & SMBO     \\ \hline
\textbf{NASNet-A\cite{NasNet}}                & 2.65 & 3.3  & 2000  & Modular & RL       \\
\textbf{pEvoNAS-C10A\cite{NASPG}}            & 2.48 & 3.6  & 1.20  & Modular & EA       \\
\textbf{DPP-Net\cite{DPPNet}}                 & 5.84 & 0.45 & 2     & Modular & SMBO     \\
\textbf{DARTS\cite{liu_darts}}                   & 2.76 & 3.3  & 4     & Modular & Gradient \\
\textbf{GDAS\cite{GDAS}}                    & 2.82 & 2.5  & 0.17  & Modular & Gradient \\
\textbf{AGNAS\cite{AGNAS}}                   & 2.46 & 3.6  & 0.4   & Modular & Gradient \\ 
\textbf{NAS-Bench-201 (Best) \cite{NASBench201}}                    & 5.63 & 1.1  & -  & Modular & - \\ \hline
 \textbf{SGAS\cite{SGAS} }               & $2.66\pm0.24$ & {3.7} & {0.25}   & Modular  & Gradient-Greedy          \\
\textbf{TRG-NAS\cite{CAIP23}}               & {4.00} & {0.45} & {4.5}   & Global  & Greedy          \\ 
\textbf{Random (Ours)}               & $6.95\pm2.18$ & $0.77\pm0.70$  & -   & Global &  -\\
\textbf{Ours (tiny)}               & \textbf{4.09} & \textbf{0.46} & \textbf{0.24}   & Global  & Greedy  \\
\textbf{Ours (mobile)}               & \textbf{3.17} & \textbf{2.49} & \textbf{0.43}   & Global  & Greedy  \\    \hline 
\end{tabular}}
\end{table}


\subsubsection{CIFAR Results} We now discuss our results on CIFAR-10 and CIFAR-100. CIFAR-10 is the most widely used dataset for NAS. So, although our work is related to global NAS, for the sake of completeness, we also compare against modular strategies, as shown in Table \ref{tab:C10}. Our approach with \textbf{tiny} settings achieves a 4.09\% error rate with a small 0.46M parameter model in just 0.24 GPU days. Further, our model size is roughly equal to that of DPP-Net \cite{DPPNet} and TRG-NAS (0.45M), but we achieve 1.84\% higher accuracy than the former and are 18$\times$ faster than later, respectively. Our network in \textbf{mobile} settings is only outperformed by Petridish among global methods, but our search is 11$\times$ faster. For CIFAR-100, we compare with Macro-NAS\cite{Macro}, the most closely related and recent global NAS method, and also with a widely used manually designed network ResNet\cite{ResNet}. Both of our \textbf{tiny} and \textbf{mobile} models outperform the best global NAS method and the manually designed network in terms of precision, parameter efficiency and search cost, as shown in Table \ref{tab:C100}. Moreover, our \textbf{tiny} model is 105$\times$ smaller than Macro-NAS \cite{Macro}, while being 1. 81\% more accurate and at 15$\times$ faster search. In general, our approach offers a better trade-off between automatic network design, high accuracy, lower model complexity, and faster search from 2$\times$ to 4$\times$.

\subsubsection{MNIST Results} 

\begin{figure}[t]
	\centering
\includegraphics[width=\linewidth]{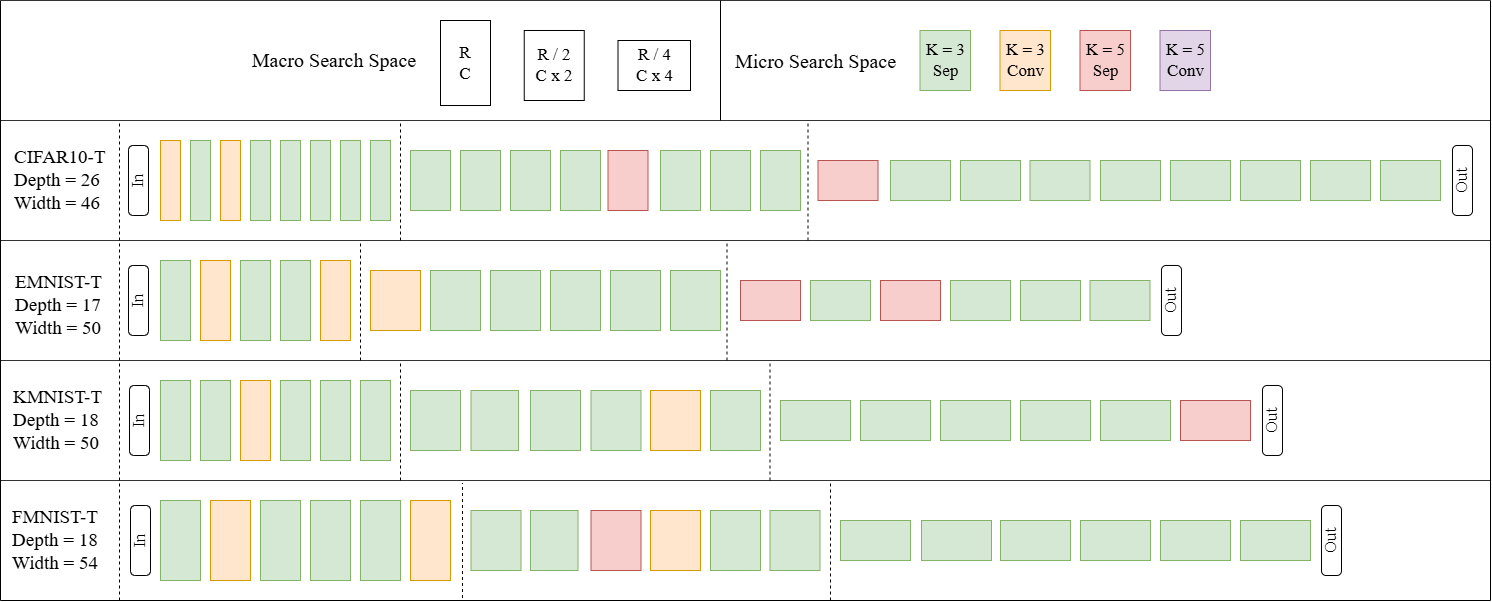}	

\caption{Tiny (T) models discovered for CIFAR-10, EMNIST, KMNIST and FashionMNIST datasets. Block height and width represents image resolution (R) and number of channels (C) respectively, while depth is represented by the total number of blocks. Different colors in Micro Search Space represent possible operation types and kernel sizes. Due to space limitations, we do not show CIFAR-100 and mobile networks.}

	\label{figure1}
\end{figure}

We achieve state-of-the-art results on EMNIST (balanced) and KMNIST datasets as compared to best manually designed networks as shown in Table \ref{tab:gray}. For \textbf{EMNIST}, our tiny model is 6$\times$ smaller than SOTA WaveMix \cite{WaveMix} (which is a highly parameter-efficient network design method), and 0.07\% better, while our mobile model achieves 0.20\% higher accuracy. For \textbf{KMNIST}, our tiny model performs 0.67\% better than the previous SOTA ViT-L/16 transformer-based network u2Net\cite{u2Net} having 300M parameters. For \textbf{FashionMNIST}, we compare our results with the state-of-the-art NAS method FT-DARTS\cite{FT-DARTS}, although it is a modular search. Our search is 9$\times$ faster and our \textbf{tiny} model is 6$\times$ smaller than FT-DARTS. Furthermore, it utilizes specialized training protocols such as drop path and random erasing. Besides this, our results with mobile network are still very competitive in terms of accuracy too without making use of advanced techniques, with the performance stemming from our proposed approach. Figure \ref{figure1} shows a few networks discovered by our framework.

\begin{table}[t]
\small
\centering
\caption{Comparison with state-of-the-art architectures for CIFAR-100}
\label{tab:C100}

\begin{tabular}{l|ccccc}
\hline
\textbf{Method} &
  \textbf{\begin{tabular}[c]{@{}c@{}}Test Err.\\ (\%)\end{tabular}} &
  \textbf{\begin{tabular}[c]{@{}c@{}}Params.\\ (M)\end{tabular}} &
  \textbf{\begin{tabular}[c]{@{}c@{}}Search Cost\\ (GPU-days)\end{tabular}} &
  \textbf{\begin{tabular}[c]{@{}c@{}}Search\\ Space\end{tabular}} &
  \multicolumn{1}{c}{\textbf{\begin{tabular}[c]{@{}c@{}}Search\\ Strategy\end{tabular}}} \\ \hline
\textbf{ResNet\cite{ResNet}}       & 29.14          & 1.7           & manual             & -      & -      \\
\textbf{Macro-NAS\cite{Macro-Neural}}    & 25.45          & 51.5          & 2.7           & Global & EA     \\
\textbf{Ours (tiny)}  & \textbf{23.64} & \textbf{0.49} & \textbf{0.17} & Global & Greedy \\
\textbf{Ours (mobile)} & \textbf{22.04} & \textbf{2.49} & \textbf{0.33} & Global & Greedy \\ \hline
\end{tabular}%

\end{table}

\begin{table}[t]
\centering
\caption{Comparison with state-of-the-art for FashionMNIST, EMNIST, and KMNIST}
\label{tab:gray}
\resizebox{\textwidth}{!}{
\begin{tabular}{@{}l|ccccccccc@{}}
\hline
 & \multicolumn{3}{c|}{\textbf{EMNIST}} & \multicolumn{3}{c|}{\textbf{KMNIST}} & \multicolumn{3}{c}{\textbf{FashionMNIST}} \\
\multirow{2}{*}{\textbf{Method}} & \textbf{\begin{tabular}[c]{@{}c@{}}Acc.\\ (\%)\end{tabular}} & \textbf{\begin{tabular}[c]{@{}c@{}}Params\\ (M)\end{tabular}} & \multicolumn{1}{c|}{\textbf{\begin{tabular}[c]{@{}c@{}}Cost\\ (Days)\end{tabular}}} & \textbf{\begin{tabular}[c]{@{}c@{}}Acc.\\ (\%)\end{tabular}} & \textbf{\begin{tabular}[c]{@{}c@{}}Params\\ (M)\end{tabular}} & \multicolumn{1}{c|}{\textbf{\begin{tabular}[c]{@{}c@{}}Cost\\ (Days)\end{tabular}}} & \textbf{\begin{tabular}[c]{@{}c@{}}Acc.\\ (\%)\end{tabular}} & \textbf{\begin{tabular}[c]{@{}c@{}}Params\\ (M)\end{tabular}} & \textbf{\begin{tabular}[c]{@{}c@{}}Cost\\ (Days)\end{tabular}} \\ \hline
\textbf{WaveMix\cite{WaveMix}} & 91.06 & 2.42 & manual & - & - & - & - & - & - \\
\textbf{u2Net\cite{u2Net}} & - & - & - & 98.68 & 300 & manual & - & - & - \\
\textbf{FT-DARTS\cite{FT-DARTS}} & - & - & - & - & - & - & \textbf{96.91} & 3.2 & 1 \\
\textbf{Ours (tiny)} & \textbf{91.20} & \textbf{0.40} & \textbf{0.36} & \textbf{99.35} & \textbf{0.42} & 0.36 & 95.90 & \textbf{0.50} & \textbf{0.11} \\
\textbf{Ours (mobile)} & \textbf{91.48} & \textbf{2.25} & \textbf{0.94} & \textbf{99.29} & \textbf{2.71} & 0.60 & 96.22 & 6.15 & \textbf{0.13} \\ \hline
\end{tabular}}
\end{table}

\subsubsection{Ablation Studies} As pointed by \cite{frustrating}, even randomly sampled architectures perform well in modular search spaces. Therefore, it cannot be determined whether the better discovered networks are due to expertly crafted search spaces or due to superior search strategies. To effectively single out the contribution of each NAS component, \cite{frustrating} suggests comparing against a simple baseline of \textbf{\textit{Randomly Sampled}} architectures and use \textbf{\textit{Relative Improvement}} metric (RI). Therefore, to show the effectiveness of our search strategy, we compare it with 10 randomly sampled architectures for CIFAR-10. In Table \ref{tab:C10}, we show that our approach achieves 2.86\% less error with 31\% fewer parameters on average. This clearly singles out the contribution of our algorithm. In addition, we use the RI metric, which is $RI = 100 \times (Acc_m - Acc_r) / Acc_r$, where $Acc_m$ and $Acc_r$ represent the accuracy of the search method and the average accuracy of the architecture sampled randomly, respectively. According to \cite{frustrating}, a good search strategy should achieve $RI > 0 $ across different runs. Our method consistently achieves an $RI > 2$ in 5 different search runs for CIFAR-10. We also test our search space with and without operations and run search for both scenarios. We used 10 different seeds for each scenario and trained candidates for 20 epochs for faster search. Searching with operations, on average, yields 0.79\% higher mean accuracy than searching without operations, that is, 89. 92\% and 89. 13\%, respectively. When searched for 600 epochs, the resulting best model without operations achieves an accuracy of 95.82\% as compared to 96\% with operations, therefore showing the contribution of operations in the search space.

\subsection{Face Recognition Experiments and Results}
\label{Recognition}

\subsubsection{Datasets} For search and training, we use the CASIA-WebFace dataset \cite{CASIA-WebFace}. For evaluation, we use three popular categories of test datasets; \textbf{1. High Quality} LFW \cite{LFW}, CFP-FP \cite{CFPFP}, CPLFW \cite{CPLFW}, AgeDB \cite{AgeDB} and CALFW \cite{CALFW}, \textbf{2. Mixed Quality} IJB-B and IJB-C \cite{IJBB,IJBC}, and \textbf{3. Low Quality} \cite{TinyFace}.

\subsubsection{Search and Training Settings} We use the same ranking mechanism and search algorithm presented in Sections \ref{perfeval} and \ref{OurAlgorithm}, respectively. We use the exact search settings as mentioned in Section \ref{ClassificationSet}, with the exception of $W_{max}$=32 for \textbf{small} and $W_{max}$=64 for \textbf{medium} settings. However, for a fair comparison with Adaface \cite{Adaface}, we slightly modify our search space to use an architecture based on inverted ResNet \cite{ResNet}. Note that the fundamental search variables are the same, that is, depth, width, operations, and kernels as introduced in Section \ref{SSDesign}. This further validates the transferability of our framework to applications beyond theory. However, due to a large number of training samples i.e. 0.5M and higher spatial resolution of 112x112, the search takes around 4 and 6 GPU days, for our small and medium models respectively, hence we early stop at best networks discovered by macro search only. For training, we follow the same protocol as in \cite{Adaface}.

\begin{table}[t]
\centering
\caption{1:1 verification accuracy and model parameters (in million; M) comparison of Adaface and our searched models on high quality datasets.}
\label{HighQuality}
\resizebox{\textwidth}{!}{%
\begin{tabular}{@{}c|c|c|cccccc@{}}
\hline
\multirow{2}{*}{Method} & \multirow{2}{*}{Model} & \multirow{2}{*}{\begin{tabular}[c]{@{}c@{}}Params.\\ (M)\end{tabular}} & \multicolumn{6}{c}{High Quality Test Data} \\  \cline{4-9}
 &  &  & LFW\cite{LFW} & CFP-FP\cite{CFPFP} & CPLFW\cite{CPLFW} & AgeDB\cite{AgeDB} & CALFW\cite{CALFW} & AVG \\ \hline
Adaface\cite{Adaface} & R-18-512 & 29.4 & 99.10 & 92.91 & 85.97 & 92.93 & 92.67 & 92.72 \\
\textbf{Ours(small)} & R-58-256 & \textbf{14.6} & \textbf{99.33} & \textbf{93.80} & \textbf{88.60} & \textbf{93.43} & \textbf{92.93} & \textbf{93.62} \\ \hline
Adaface\cite{Adaface} & R-50-512 & 49.0 & \textbf{99.40} & 94.71 & 89.33 & 93.88 & 93.33 & 94.13 \\
\textbf{Ours(medium)} & R-34-512 & \textbf{37.1} & 99.35 & \textbf{95.13} & \textbf{89.83} & \textbf{94.27} & \textbf{93.68} & \textbf{94.45} \\ \hline
\end{tabular}
}
\end{table}

\subsubsection{High Quality Results} 
\begin{figure}[t]
\includegraphics[width=\linewidth]{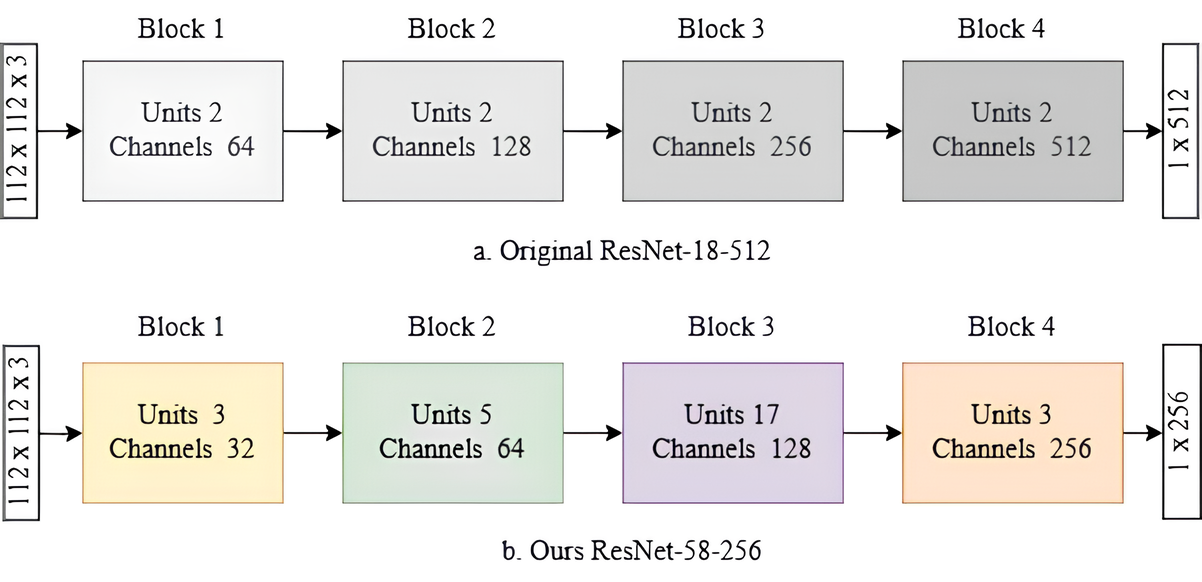}	

\caption{Architectural comparison of a. R-18-512 with our b. R-58-256. }

	\label{figure2}
\end{figure}

For small settings, we discover a model with 58 layers but half the number of channels used by \cite{Adaface}, that is, 256 in final block compared to 512. Our model performs better on all high-quality datasets for the 1:1 verification task, as shown in Table \ref{HighQuality}. A comparison of our small model and ResNet-18 is shown in Figure \ref{figure2}. Ours has 2$\times$ fewer parameters than ResNet-18. For medium settings, we discover a model with just 34 layers, having 1.3$\times$ fewer parameters as compared to ResNet-50, but still performing better on all datasets except LFW. 

\begin{table}[t]
\centering
\caption{For mixed quality datasets IJB-B and IJB-C, TAR@FAR=0.01\% are reported. For low quality dataset TinyFace, Top-1, Top-5 and Top-20 rank retrieval is used.}
\label{Mixed and Low Quality}
\resizebox{\textwidth}{!}{%
\begin{tabular}{c|c|c|cc|ccc}
\hline
\multirow{3}{*}{Method} & \multirow{3}{*}{Model} & \multirow{3}{*}{\begin{tabular}[c]{@{}c@{}}Params\\ (M)\end{tabular}} & \multicolumn{2}{c|}{Mixed Quality} & \multicolumn{3}{c}{Low Quality} \\ \cline{4-8} 
 &  &  & \multicolumn{1}{c|}{IJB-B\cite{IJBB}} & IJB-C\cite{IJBC} & \multicolumn{3}{c}{TinyFace\cite{TinyFace}} \\
 &  &  & \multicolumn{1}{c|}{TAR@FAR=0.01\%} & TAR@FAR=0.01\% & Rank-1 & Rank-5 & Rank-20 \\ \hline
Adaface\cite{Adaface} & R-18-512 & 29.4 & \textbf{93.11} & \textbf{94.03} & 54.85 & 60.30 & 64.10 \\
\textbf{Ours(small)} & R-58-256 & \textbf{14.6} & 92.06 & 92.60 & \textbf{57.78} & \textbf{62.66} & \textbf{66.36} \\ \hline
Adaface\cite{Adaface} & R-50-512 & 49.0 & \textbf{94.52} & \textbf{95.34} & 57.32 & 62.63 & 66.66 \\
\textbf{Ours(medium)} & R-34-512 & \textbf{37.1} & 93.22 & 93.65 & \textbf{58.66} & \textbf{64.54} & \textbf{67.97} \\ \hline
\end{tabular}
}
\end{table}
\subsubsection{Mixed and Low Quality Results} Table \ref{Mixed and Low Quality} shows our results on mixed and low-quality test datasets. Given that our models are much smaller, we achieve competitive results on both IJB-B and IJB-C datasets \cite{IJBB,IJBC}. On Tinyface \cite{TinyFace}, both of our models achieve better accuracy in the rank-1, rank-5, and rank-20 identification task.

\subsubsection{Limitations and Future Work}
Although, we have used face recognition to demonstrate real-world applications of our framework, in general, given enough computational resources, it can yield accurate and parameter-efficient models for other computer vision applications too. Moreover, since we are searching on the whole dataset to evaluate candidates, the dataset size and image resolution influence candidates' training time, and hence search speeds. Therefore, further speedups can be investigated in the context of larger datasets such as ImageNet \cite{ImageNet} and CASIA-Webface \cite{CASIA-WebFace} leveraging data subsets instead of entire data. Moreover, we observe that macro search coupled with good training protocols achieves more significant accuracy gains compared to micro search. Therefore, another promising research direction is to redesign the framework for joint macro architecture and training protocol search to improve overall accuracy and search efficiency.

\section{Conclusion}

Contrary to trending modular search that offers partial network discovery, we revisit NAS for end-to-end network discovery. Our search space offers networks of varying complexity and is designed such that it can be efficiently navigated by the proposed search strategy. Our architecture aware ranking mechanism leads to much faster search as compared to existing methods, yet consistently assists in discovering highly accurate tiny and mobile architectures. Lastly, we demonstrate the transferability of our framework from theory to real-world applications of face verification and identification with the continued trend of discovering more accurate, yet smaller networks. 



\subsubsection{Competing Interests} The authors declare that they have no competing interests.

\subsubsection{Funding Information} This work was supported by the European Union\textquotesingle Horizon 2020 research and innovation program under grant agreement No 739551 (KIOS CoE) and from the Government of the Republic of Cyprus through the Directorate General for European Programs, Coordination and Development.

\subsubsection{Author Contributions} Shahid Siddiqui designed the NAS framework, conducted the experiments, and wrote the initial draft of this manuscript. Christos Kyrkou suggested experiments to test the framework and further improved the draft. Theocharis Theocharides provided guidance for the final manuscript.

\subsubsection{Data Availability Statement} All search and training codes, as well as logs are available at: \href{https://github.com/siddikui/Efficient-Macro-Micro-NAS}{https://github.com/siddikui/Efficient-Macro-Micro-NAS}

\subsubsection{Research Involving Humans and/or Animals} Not applicable.

\subsubsection{Informed Consent} Informed consent was obtained from all participants included in the study.


\end{document}